\documentclass[sigconf, nonacm]{acmart} 
\renewcommand\footnotetextcopyrightpermission[1]{}
\usepackage{hyperref}
\usepackage{url}
\usepackage{bbm} 
\usepackage{algorithm}
\usepackage{makecell}

\usepackage[utf8]{inputenc}
\usepackage{enumitem}

\usepackage{multirow}

\usepackage{makecell} % for better cell formatting
\usepackage{xcolor}   % for colors
\usepackage{colortbl} % optional, for background colors

\graphicspath{ {./figures/} }

\begin{document}

%%
%% The "title" command has an optional parameter,
%% allowing the author to define a "short title" to be used in page headers.
\newcommand{\pipeline}{LongRecall}
\title{LongRecall: A Structured Approach for Robust Recall Evaluation in Long-Form Text}

%%
%% The "author" command and its associated commands are used to define
%% the authors and their affiliations.
%% Of note is the shared affiliation of the first two authors, and the
%% "authornote" and "authornotemark" commands
%% used to denote shared contribution to the research.
\author{MohamamdJavad Ardestani}
\email{ardestan@ualberta.ca}
\affiliation{%
  \institution{Department of Computing Science University of Alberta}
  \city{Edmonton}
  \state{Alberta}
  \country{Canada}
}

\author{Ehsan Kamalloo}
\email{ehsan@servicenow.com}
\affiliation{%
  \institution{ServiceNow Research}
  \city{Montreal}
  \state{Quebec}
  \country{Canada}
}

\author{Davood Rafiei}
\email{drafiei@ualberta.ca}
\affiliation{%
  \institution{Department of Computing Science University of Alberta}
  \city{Edmonton}
  \state{Alberta}
  \country{Canada}
}

%%
%% The abstract is a short summary of the work to be presented in the
%% article.
\begin{abstract}
The completeness of machine-generated text---ensuring that it captures all relevant information---is crucial in domains such as medicine and law and in tasks like list-based question answering (QA), where omissions can have serious consequences. However, existing recall metrics often depend on lexical overlap, leading to errors with unsubstantiated entities and paraphrased answers, while LLM-as-a-Judge methods with long \emph{holistic} prompts capture broader semantics but remain prone to misalignment and hallucinations without structured verification.
We introduce \textbf{{\pipeline}}, a general three-stage recall evaluation framework that decomposes answers into self-contained facts, successively narrows plausible candidate matches through lexical and semantic filtering, and verifies their alignment through structured entailment checks. This design reduces false positives and false negatives while accommodating diverse phrasings and contextual variations, serving as a foundational building block for systematic recall assessment.
We evaluate {\pipeline} on three challenging long-form QA benchmarks using both human annotations and LLM-based judges, demonstrating substantial improvements in recall accuracy over strong lexical and LLM-as-a-Judge baselines.
All code, datasets, generated content, prompts, and human annotation instructions are available in our anonymized repository, which will be made public upon the paper’s acceptance and publication.
\end{abstract}

%% Keywords. The author(s) should pick words that accurately describe
%% the work being presented. Separate the keywords with commas.
\keywords{Evaluation Methodologies, Long-Form Text Evaluation, Long-Form Question Answering, Generative Question Answering}

\maketitle

\section{Introduction}

\begin{table*}[t]

\centering
{ 

\renewcommand{\arraystretch}{1.2}
  
% \begin{tabularx}{\textwidth}{|X|}
\begin{tabular}{|p{\dimexpr\textwidth-2\tabcolsep-2\arrayrulewidth\relax}|}
\hline
\textbf{Question}: Which European film directors influenced European cinema between 1930 and 1970?\\
\hline
\textbf{Reference Answer} \\
\hline

\textcolor{blue}{[Federico Fellini]}: He pioneered blending fantasy and autobiography in European cinema, directing the influential film 8½ (1963).

\textcolor{olive}{[Jean-Luc Godard]}: Revolutionized European cinema with \emph{À bout de souffle} (1960), became a hallmark of the French New Wave movement. 

\textcolor{orange}{[Bergman]}: Ingmar Bergman was a master of existential European cinema in the 1960s, crafting deeply introspective narratives.

\textcolor{purple}{[Ryszard Boleslawski]}: Directed \emph{Les Misérables}, shaping early European cinema with his sophisticated visual storytelling.\\

\hline
\textbf{Generated Answer} \\
\hline

The film Eight and a Half, directed by the Italian \textcolor{blue}{II Maestro} in 1960s, became renowned for merging imagination with personal experience on screen, fundamentally shaping the future of cinema in Europe ... .
\textcolor{purple}{Richard Boleslavsky}'s work, especially the 1935 adaptation of \emph{Les Misérables}, has influenced many but is often overlooked in discussions of European film ... .
Vincenzo Caputo, a contemporary director, draws inspiration from the visual motifs seen in the films of 
\textcolor{olive}{Jean-Luc Godard} ... .
\textcolor{orange}{Bergman Jr.} Ingmar's son, represents a contemporary wave of stylists reshaping modern European art cinema ... .\\
\hline
\end{tabular}
}

\caption{\label{tab:lexical_failure_case}
Failure modes of lexical recall metrics: \textbf{False Negatives (FN)} arise when valid references are missed due to paraphrasing or alternative names, such as ``\textcolor{blue}{II Maestro}'' for \textcolor{blue}{Federico Fellini} or \textcolor{purple}{Richard Boleslavsky} for \textcolor{purple}{Ryszard Boleslawski}. \textbf{False Positives (FP)} occur when surface matches lack sufficient evidence, such as matching \textcolor{orange}{Bergman Jr.}, who refers to a different person than \textcolor{orange}{[Bergman]}, or when \textcolor{olive}{Jean-Luc Godard} appears in the generated answer solely as part of the evidence or background for another director (Vincenzo Caputo), rather than being presented as a distinct answer director with its own justification.}

\end{table*}

% Problem and challenges
As large language models (LLMs) \citep{brown2020language,achiam2023gpt,grattafiori2024llama,yang2024qwen2,groeneveld-etal-2024-olmo,liu2024deepseek} become increasingly prevalent in text generation tasks---ranging from report writing~\citep{du-etal-2024-llms,agarwal2024llms,liang2025widespread} to complex question answering (QA)~\citep{hendrycks2021measuring,rein2024gpqa,wang2024mmlupro}---ensuring the factual accuracy and completeness of long-form responses remains crucial~\citep{dmonte2024claim}. This is especially critical in high-stakes domains such as medicine~\citep{zhou2023synthetic} and law~\citep{magesh2024hallucination}, where missing even a single key fact or piece of evidence may incur serious real-world consequences~\citep{bommasani2021opportunities,weidinger2021ethical,wachter2024large}.

The primary challenge in measuring factuality arises from lexical variations in how the identical content may be expressed with different wording, paraphrases, or sentence structures, and, in some cases, entailment relationships that require reasoning beyond surface-level text matching.
For example, consider the responses in Table~\ref{tab:lexical_failure_case} for the question: \textit{Which European film directors influenced European cinema between 1930 and 1970?} The generated answer includes only the nickname for \textit{Federico Fellini} and the English transliteration of \textit{Ryszard Bolesławski}. The name \textit{Bergman} appears in both texts but refers to different individuals, further complicating interpretation. Additionally, the supporting evidence for entities such as \textit{Jean-Luc Godard} is missing from the generated response. These discrepancies make it  difficult to determine whether a generated response fully captures all the evidence provided in a reference text.

% Existing solutions and their limitations 
Existing evaluation methods for generative QA models primarily emphasize precision-oriented metrics, such as factual accuracy~\citep{factscore,wei2024longform,song-etal-2024-veriscore,tang-etal-2024-minicheck} and hallucination detection~\citep{manakul-etal-2023-selfcheckgpt,li-etal-2023-halueval,dhuliawala-etal-2024-chain,farquhar2024detecting,sui-etal-2024-confabulation}, largely neglecting systematic measures of recall. However, recall is crucial for assessing whether all key points are included and adequately supported, not merely whether the presented information is correct. 
Despite its importance in evaluating long-form answers, recall assessment remains limited and less explored in general. Existing approaches \citep{QAMPARI} primarily measure recall based on surface-form lexical matching.

However, these metrics break down when faced with the expressive variation of modern LLMs \citep{kamalloo-etal-2023-evaluating}: they produce false negatives when valid answers are phrased using synonyms, paraphrases, nicknames, or alternative spellings, and false positives when surface-form overlap occurs despite missing essential details. Table~\ref{tab:lexical_failure_case} illustrates these failure modes, demonstrating how lexical-based metrics fail to detect valid responses due to insufficient lexical overlap.

% General version
Alternatively, holistic prompting methods (``LLM-as-a-Judge'') evaluate answer completeness by processing the entire context within an LLM simultaneously~\citep{zheng2023judging}. While effectively capturing broader semantic similarities and implicitly handling paraphrases, they face significant practical issues~\citep{li2024generation,zhu2025judgelm,li2025preference}. For example, the evaluator LLM frequently misaligns reference segments by mapping them onto themselves rather than correctly matching them to generated counterparts. Moreover, as  the reference and generated answers grow longer, input complexity makes it harder for the LLM to track essential details, often causing it to overlook subtle yet critical nuances. Such problems are particularly prevalent when different parts of the answers share similar contexts, leading to inconsistent evaluations, and hallucinations.

To address this challenge, we propose \textbf{LongRecall}, a general three-stage framework for robust recall assessment in long-form QA, designed as a reusable building block for diverse text generation tasks. LongRecall systematically verifies whether generated outputs adequately cover all required information, an essential requirement in many real-world scenarios, including \textit{retrieval-augmented generation} (ensuring answers are grounded in a reference corpus), \textit{multi-document summarization} (capturing all salient facts from source documents), \textit{factoid and biomedical QA} (measuring completeness against authoritative sources), and \textit{model comparison} (assessing smaller models against outputs from larger, stronger systems).

Our modular approach aims to achieve accuracy, interpretability, and scalability through three complementary stages. First, \textit{fact extraction} decomposes long-form answers into self-contained facts. Next, \textit{candidate selection} identifies potential matches between extracted reference and generated facts using a two-step similarity matching---first lexical similarity to retrieve close surface-level matches, followed by semantic similarity to account for paraphrasing and diverse expressions. Finally, \textit{entailment checking} utilizes a structured, LLM-based verification approach that correlates with human judgment to determine whether selected candidates fully entail their reference counterparts. Through extensive experimentation on three challenging long-form QA benchmarks, QAMPARI~\citep{QAMPARI}, RoMQA~\citep{RoMQA}, and ExpertQA~\citep{expertqa}, we demonstrate significant improvements in recall accuracy compared to existing lexical and holistic evaluation methods, across diverse answer structures.

\section{Related Work}

\paragraph{Long-form Text Evaluation:}
Traditionally, the quality of long-form text---ranging from paragraphs to articles---were assessed using word overlap metrics such as BLEU~\citep{bleu} for machine translation and ROUGE~\citep{lin-2004-rouge} for abstractive summarization.
However, these metrics fail to capture deeper semantics beyond surface-level features \citep{liu-etal-2016-evaluate,novikova-etal-2017-need}.
As a remedy, LM-based metrics including BERTScore~\citep{bertscore}, BLEURT~\citep{bleurt}, BARTScore~\citep{bartscore}, and AlignScore~\citep{zha-etal-2023-alignscore} have sought to incorporate semantics by measuring similarity within an LM representation space~\citep{jiang2019semantic}.
% Yet, human evaluation remains the most reliable method for evaluating long text despite being time-consuming and costly. 
The recent success of LLMs has made them a strong alternative to humans for this purpose \citep{xu-etal-2023-instructscore,kim2024prometheus,kim-etal-2024-prometheus}. 
Numerous studies have explored using LLMs for automated evaluation of long text in Question Answering~\citep{kamalloo-etal-2023-evaluating} and Retrieval-Augmented Generation~\citep{saad-falcon-etal-2024-ares,es-etal-2024-ragas,peng2024unanswerability}.
Our work is in line with this body of work, as we also leverage LLMs to evaluate long-form text. 
But, instead of fine-tuning or crafting long prompts, our approach follows a structured pipeline design.

\begin{table*}[t]
\centering
{ 
% \fontsize{10}{12}\selectfont 
\renewcommand{\arraystretch}{1.4}
\begin{tabular}{|p{\dimexpr\textwidth-2\tabcolsep-2\arrayrulewidth\relax}|}
\hline

\hline
\textbf{Extracted Facts from Generated Answer} \\
\hline

[\textcolor{blue}{{II Maestro}}]: The film Eight and a Half, directed by the Italian II Maestro in 1960s, became renowned for merging imagination with ... .

[\textcolor{purple}{Richard Boleslavsky}]: Richard Boleslavsky's work, especially the 1935 adaptation of \emph{Les Misérables}, has influenced many but is often ... .

[Vincenzo Caputo]: Vincenzo Caputo, a contemporary director, draws inspiration from the visual motifs seen in the films of ... .

[\textcolor{orange}{Bergman Jr.}]: Bergman Jr., Ingmar's son, represents a contemporary wave of stylists reshaping modern European art cinema ... .\\

\hline
\end{tabular}
}
\caption{
Extracted Facts from Generated Answer in Table~\ref{tab:lexical_failure_case}.}

\label{tab:fact_extraction}
\end{table*}

\paragraph{Factuality Evaluation:}
Machine-generated text is often prone to factual errors~\citep{maynez-etal-2020-faithfulness,raunak-etal-2021-curious,dziri-etal-2022-origin,mallen-etal-2023-trust}, which erodes trust in its use for real-world applications \citep{ji2023survey}. As a result, automatic evaluation of factuality in long-form text is a critical task \citep{rashkin2021measuring,dziri-etal-2022-faithdial,kamalloo2023hagrid}.
Many existing methods \citep{factscore,wei2024longform,song-etal-2024-veriscore} decompose long text into simpler statements and verify factual accuracy using entailment or external tools (e.g., search engines).
Other works generate synthetic data from a strong LLM to fine-tune a smaller LM to be a critic \citep{vu-etal-2024-foundational,tang-etal-2024-minicheck}.
However, this line of work focuses primarily on factual precision.
In this paper, we aim to measure factuality recall to ensure machine-generated text captures \emph{all relevant facts} from an input corpus.
% Recall is particularly crucial in high-stake domains such as law and medicine.
% TODO: ERecall and ARecall in QAMPARI

\section{Problem Formulation and Challenges}
\label{problem-formulation}

Let \( F \) denote a reference set of facts and \( G \) a set of model-generated facts, both typically extracted from free-form text. Our goal is to measure the \textit{recall} of \( G \) with respect to \( F \)—that is, the proportion of facts in \( F \) that are \textit{covered} or entailed by one or more facts in \( G \).

For a given reference fact \( f \in F \), let \( C(f, G) \subseteq G \) denote the set of generated facts that cover or entail \( f \). This set may be empty if no fact or combination of facts in \( G \) sufficiently covers \( f \). Using this, we define recall as:
\begin{equation}
R(G) = \frac{1}{|F|} \sum_{f \in F} \mathbbm{1}_{\{C(f,G) \neq \emptyset\}}
\label{eq:recall}
\end{equation}

Here, \( \mathbbm{1}_{\{C(f,G) \neq \emptyset\}} \) is an indicator function that returns 1 if \( C(f,G) \) is non-empty---that is, if there exists a subset of \( G \) that covers \( f \)---and 0 otherwise. When evaluated over a dataset, recall is computed as the mean of \( R(G) \) across all instances.

Evaluating recall for long-form text introduces several challenges. First, the facts in \( F \) and \( G \) are often expressed as complex, free-form natural language statements rather than being decomposed into simple, self-contained facts. This makes it difficult to directly verify entailment. Second, for each fact \( f \in F \), every subset of \( G \) is a potential candidate to support or entail \( f \). The number of such candidates is given by the powerset of \( G \), which grows exponentially with the size of \( G \), making exhaustive search computationally infeasible. Third, even when candidate sets are identified, determining whether a fact (or set of facts) entails another is non-trivial, typically requiring sophisticated natural language inference models, as previously noted in work on textual entailment~\cite{dagan2005pascal}.

\begin{figure*}[t]
  \includegraphics[width=\linewidth]{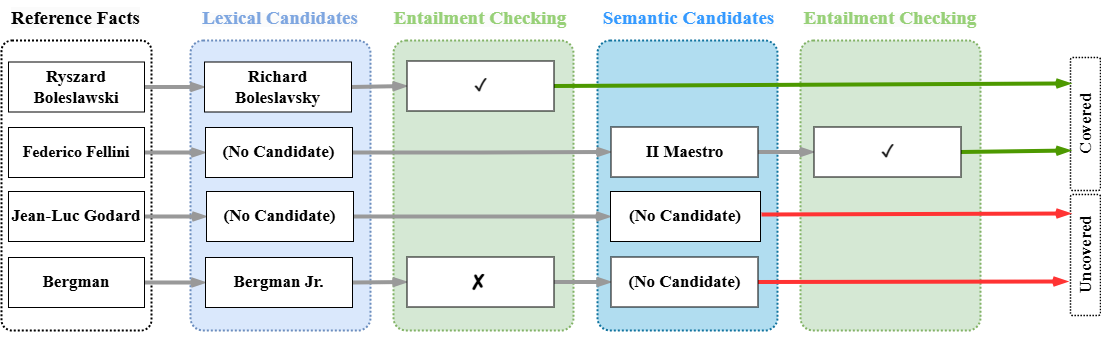} 
    \caption{\label{fig:pipeline_diagram} Pipeline stages are illustrated using the example in Table~\ref{tab:lexical_failure_case}. For illustration, only anchor spans are shown; full facts are used in the actual pipeline. Candidates are filtered by lexical similarity and verified via entailment checks. If still unverified, semantic filtering and a second entailment check are applied; any fact not confirmed by entailment is marked as uncovered.}
\end{figure*}

\section{Approach}

Our approach introduces a modular, extensible framework for evaluating the completeness of long-form answers through a structured three-stage pipeline---fact extraction, candidate selection, and entailment checking---designed to adapt across different domains and answer structures while maintaining interpretability.

\subsection{Fact Extraction}

To enable fine-grained recall evaluation, long-form answers must be decomposed into self-contained, verifiable facts \cite{zhang-bansal-2021-finding,factscore}. A fact is \emph{self-contained} if it can be understood without relying on surrounding context. A self-contained fact is not necessarily atomic and can be a compound. The level of granularity chosen for this decomposition directly impacts the reliability of recall metrics.

Since our goal is to match generated text against a reference set, it is essential that both sets follow the same level of granularity to ensure consistency in evaluation. Our fact extraction strategy is therefore guided by alignment with the structure of the reference. Specifically, when the reference is structured as an itemized list, common in list-based QA, we extract each list entry as a separate fact using lightweight, rule-based heuristics that rely on common delimiters such as numbers, bullets, colons, and parentheses. In cases where the reference is presented as free-form text but annotated at the sentence or clause level, we mirror the same segmentation to the generated answer, treating each resulting segment as a standalone fact. 
In cases where reference annotations do not include list formatting or clause-level annotation, semantic decomposition methods such as FActScore~\citep{factscore} can be employed to consistently segment both reference and generated text into atomic facts using LLMs.

We distinguish between two common types of extracted facts in long-form QA: \emph{split-span} and \emph{monolithic} facts. Split-span facts (as included in Table~\ref{tab:fact_extraction}) consist of a short anchor, such as an entity or answer phrase, paired with a supporting evidence span. Monolithic facts, by contrast, are expressed as single statements that contain all information in one span. Our pipeline is designed to support both types, ensuring consistent granularity and comprehensive coverage in recall evaluation.

To ensure semantic independence, we apply a one-shot coreference resolution prompt inspired by~\citet{le2023large}. This step rewrites anaphoric expressions (e.g., pronouns or definite noun phrases) to include explicit mentions, thereby eliminating cross-sentence dependencies and preserving the self-contained nature of each fact. Table~\ref{tab:fact_extraction} presents the results of applying fact extraction to the model-generated text from Table~\ref{tab:lexical_failure_case}.

\subsection{Candidate Selection}

To reduce the computational cost of exhaustively comparing each reference fact $f \in F$ against all extracted facts from generated answer in $G$, we introduce a candidate selection module. For each reference fact, this module uses lexical and semantic filters to identify a focused subset of $G$ that is most likely to entail the reference, ensuring that all potentially relevant matches are retained for entailment checking.

\paragraph{Lexical Filtering}
Lexical filtering provides a high-precision and fast signal by exploiting surface-form similarity between facts. We implement this using \textbf{Fuzzy Jaccard Similarity}~\citep{jaccard}, which measures token-level overlap while tolerating minor spelling variations. To break ties and refine rankings, we apply \textbf{Longest Common Subsequence (LCS)}~\citep{lcs}, which captures fine-grained character-level overlap. For each reference fact, we select the top-$k$ generated candidates that exceed a dataset-specific similarity threshold.

\paragraph{Semantic Filtering} 
While lexical filtering offers high precision, it fails to capture matches involving abstraction or paraphrasing---for example, missing alignments between ``\textit{car}'' and ``\textit{automobile}.'' To overcome this, we incorporate semantic filtering using dense vector representations and cosine similarity to estimate the semantic closeness between facts. For each reference fact, we rank generated candidates by similarity and retain the top-$k$ entries that exceed a predefined threshold. This approach tends to retrieve paraphrased or conceptually similar content that lacks surface-form overlap, but its broader matching scope can lead to false positives, such as aligning ``\textit{Nobel Prize in Literature}'' with ``\textit{Nobel Peace Prize}.'' 

By combining the precision of lexical filtering with the flexibility of semantic matching, we aim to construct a more balanced candidate set---improving recall of paraphrased facts without compromising accuracy. This complementary behavior is especially valuable given the structural diversity of extracted facts: lexical signals perform well on \textbf{split-span facts}, where entity anchors provide strong alignment cues, while semantic similarity excels on \textbf{monolithic facts}, which often exhibit limited surface overlap. Our sequential pipeline leverages this division to support any format of free-form answer efficiently.

\subsection{Entailment Checking}

While candidate selection identifies potentially relevant facts, it does not guarantee that those candidates entail the reference facts. Using a short, illustrative sample for clarity, consider the question ``\textit{Who directed Inception?}'' and the reference fact ``\textit{Christopher Nolan directed the 2010 film Inception},'' lexical and semantic filters might admit candidates such as ``\textit{Christopher Nolan is a filmmaker}'' or ``\textit{Tenet was directed by Nolan.}'' These are topically related but do not entail the reference fact. In contrast, ``\textit{The 2010 film Inception was directed by Christopher Nolan}'' would be a valid entailment. Our entailment checking module is designed to make this distinction---accepting only candidates that truly entail the reference.

We employ LLMs with two complementary prompting strategies to verify entailment: \textit{one-to-one} verification and \textit{multiple-choice} entailment. 

\paragraph{One-to-One Entailment} This strategy is employed when a single strong candidate is identified for a reference fact. The LLM is prompted to assess the candidate-reference pair and classify the relationship as either \textit{``Match''} or \textit{``No Match''}. This setting offers high precision with minimal computational cost. 

\paragraph{Multiple-Choice Entailment}
When several candidates may jointly entail a reference fact, or when no single candidate suffices, we use a multi-choice prompt.
The LLM is instructed to either select the best individual candidate or identify a minimal subset of candidates that collectively entail the reference fact. This strategy supports both comparative and compositional reasoning.

Together, these two strategies offer a balance of reasoning fidelity and efficiency. The one-to-one approach ensures fine-grained entailment judgments with minimal token usage, while the multiple-choice approach reduces redundant calls by evaluating several candidates in a single inference. A reference fact that is not entailed under either strategy is deemed \textit{not covered}. Entailment prompts are available in our code repository.

\subsection{LongRecall Pipeline}

LongRecall integrates fact extraction, candidate selection, and entailment checking into a unified, verify-as-you-go workflow. As illustrated in Figure~\ref{fig:pipeline_diagram},  the system processes each reference fact through successive filtering stages. First, lexical filtering identifies high-precision candidates based on surface-form similarity. If entailment is confirmed for any of these candidates, the reference fact is marked as \textit{covered}, and processing stops. Otherwise, blocked pairs are excluded to avoid redundancy. Uncovered facts are then passed to the semantic filtering stage, which retrieves a new candidate set---excluding previously blocked pairs and capturing paraphrased or conceptually similar candidates that lexical filtering may have missed---and performs a second round of entailment checking. If no entailment is found across both stages, the reference fact is marked as \textit{uncovered}.

The pipeline’s staged ordering is intentional: 
lexical filtering runs first due to its low computational cost and strong precision, while semantic filtering---being more resource-intensive---is reserved for paraphrastic or abstract alignments. Performing entailment checks immediately after each filtering stage allows for early retirement of covered facts, reduces the size of candidate pools, and keeps LLM prompts concise. This structure ensures that overt matches are resolved efficiently through lexical filtering, while only  subtle or abstract matches proceed to the semantic stage. By combining early exits with targeted entailment prompts, the pipeline minimizes computational overhead while maintaining high alignment fidelity.

\section{Experiments}

\subsection{Datasets}

We evaluate our framework's generalization capability across three diverse benchmarks that exemplify different contexts and content structures encountered in real-world QA applications. All datasets are publicly available under their respective licenses, and our dataset splits and preprocessing scripts are available in our anonymized GitHub repository.

\textbf{QAMPARI}~\citep{QAMPARI} is a list-based QA dataset with approximately  62K training questions, along with 1K each for validation and test. Each question is paired with an average of 13 answer items. The questions fall into Simple, Intersection, and Composition types, reflecting varying reasoning complexities. They are semi-automatically generated from Wikipedia tables and knowledge graphs, then paraphrased by humans for fluency and diversity. For each question, answers are formatted as split-span entries: a short anchor (typically an entity name) paired with a Wikipedia sentence or snippet that serves as supporting evidence. For our experiments, we use the first 500 samples from the test set, covering all three question types.

\textbf{ExpertQA}~\citep{expertqa} is a domain-specific QA dataset comprising expert-curated questions spanning 32 fields of study, with nearly 2K questions in total. Unlike the list-based datasets, answers are provided as free-form paragraphs and are segmented into sentence-level \emph{claims} representing monolithic segments, with each response containing an average of 6 claims. Each claim is annotated for both informativeness and correctness. For evaluation, we retain only those claims labeled as either ``Very relevant'' or ``A bit relevant'' (informativeness) and ``Definitely correct'' or ``Probably correct'' (correctness), following the dataset guidelines. After applying our filtering criteria, we randomly sample 200 answers for evaluation.

\textbf{RoMQA}~\citep{RoMQA} is another list-based QA dataset consisting of 28K questions with an average of 108 correct answers. RoMQA has a systematic organization: human-authored questions are grouped into clusters based on shared Wikidata ``constraints,'' such as `occupation = \textit{pianist}' or `birthplace = \textit{Paris},' that define which entities satisfy each query. This constraint-driven design enables comprehensive coverage of entity relationships while requiring reasoning over long evidence text. To reduce redundancy and ensure coverage diversity, we adapt RoMQA by selecting one random question from each cluster of semantically related queries. Answers follow the similar split-span format as QAMPARI, pairing each entity anchor with a supporting Wikipedia snippet, yielding 500 samples for our experiments.

This diversity allows us to evaluate LongRecall across fundamentally different answer structures: compact entity collections (QAMPARI), expert-authored paragraphs (ExpertQA), and large-scale constraint-driven responses (RoMQA), spanning a range of reasoning types and domain complexities.

\subsection{Experimental Setup}

\subsubsection*{Answer Generation and Model Selection}
We generate answers for RoMQA and QAMPARI using GPT-4o~\citep{hurst2024gpt}, and use Claude-3.5-Sonnet~\citep{claude} for ExpertQA. This model choice for ExpertQA avoids potential stylistic alignment with GPT-based reference answers in that dataset, ensuring model-family diversity between reference and generated answers and minimizing evaluation bias. All prompt templates used with each dataset are available in our code repository.

\subsubsection*{Data Preprocessing}
To ensure consistent evaluation, we apply a unified preprocessing pipeline to all reference and generated outputs. We first apply redundancy filtering to remove duplicate or near-duplicate spans, retaining only distinct supporting evidence. Unicode normalization standardizes symbols (e.g., curly quotes, dashes) into canonical ASCII forms. We also apply formatting regularization to unify casing, punctuation, and bracketing conventions, ensuring that minor typographic differences do not affect surface-level matching or entailment judgments. All preprocessing scripts are available in our code repository.

\subsubsection*{Hyperparameter Configuration}
To determine the appropriate thresholds and hyperparameter values for our pipeline, we manually tested a range of settings for each component on 50 samples from each benchmark dataset. For lexical filtering, we selected a Fuzzy-Jaccard threshold between 0.30-0.40 and an LCS threshold between 0.60-0.80, which consistently yielded strong performance. For semantic filtering, we use Google's Gemini embedding model with Vertex AI's task-type framework to embed reference answers as retrieval queries and generated answers as retrieval documents, with cosine similarity thresholds in the range of 0.60-0.65 proving most effective at removing spurious matches while preserving valid paraphrases. These values were empirically identified as optimal compared to both stricter and more lenient alternatives.

Empirical analysis also showed that in nearly all cases, at most three generated candidates were sufficient to cover each reference answer; instances requiring more than three were rare. All LLM calls used default parameters with temperature = 0.0 to ensure deterministic outputs. All thresholds and hyperparameters are fully configurable through modular, plug-and-play configuration files provided in our codebase. The exact values used for each benchmark, along with embedding model configurations, are included there for full transparency and reproducibility. LongRecall achieves fast processing at 3.5–5 seconds per example using online API calls in single-threaded mode, and can be further accelerated through local inference and multi-threading supported by its modular design.

\subsection{Baselines for Computing Recall}
\label{subsec:Recall Calculation Methods and Baselines}

We compare our LongRecall pipeline against established recall metrics that compute recall via an explicit mapping between reference facts and generated facts; for each question, every reference fact $f \in F$ is matched to zero or more generated facts $g \in G$, and recall is measured by aggregating binary accuracy scores across all such pairs.

\paragraph{Surface‐Level Matching:}
We adopt two lexical baselines from \citet{QAMPARI}: \textbf{(i) ARecall} that marks a reference fact ($f$) as ``covered'' only if its anchor span---after case folding and whitespace normalization---appears verbatim in any generated fact ($g$), and \textbf{(ii) ERecall} which extends ARecall by computing longest common substring (LCS) between the full text of reference $f$ and generated fact $g$, regardless of whether they are split‐span or free-form. $f$ is considered ``covered'' when LCS, normalized by length, exceeds a predefined threshold. We set the threshold to 0.4, which yielded the best performance in our trials over values ranging from 0.3 to 0.8 on a sample subset.
 %Both metrics enforce strict one‐to‐one alignments, so they do not capture paraphrases or cases where evidence is distributed across multiple generated facts.

\paragraph{Matching via LLM-as-a-Judge:} 
We incorporate a holistic ``LLM-as-a-Judge'' baseline~\citep{zheng2023judging} where an LLM is instructed to assign each reference fact \(F\) to zero or more generated facts \(G\) based on its own internal reasoning. To do so, we construct a prompt that consists of a question and its corresponding reference facts \(F\) along with generated facts \(G\) in addition to detailed instructions of the task.
This baseline leverages the LLM’s broad understanding without relying on explicit candidate selection oe entailment checking. However, diagnosing failures here is challenging, since all reasoning is embedded in a single holistic prompt, thus offering limited clues about the models' underlying reasoning. The complete prompt templates are provided in our code repository.

These two baselines represent two ends of the spectrum: shallow lexical matching versus holistic but unstructured semantic assessment. Our proposed method, LongRecall, bridges this gap by combining modular reasoning steps with structured verification, enabling more interpretable and reliable recall estimation.

\section{Evaluation}

Human evaluation is the most reliable method for assessing the effectiveness of LongRecall and its baselines, but it is prohibitively time-consuming and costly at scale.
To address this, we investigated whether LLMs can serve as a scalable proxy for human judgment by collecting human annotations on two strategically sampled sets and measuring agreement with automated LLM-based evaluations. Specifically, we sampled 80 reference–generated fact pairs from each dataset to ensure coverage of key fact-structure types across our benchmarks: QAMPARI (split-span dataset) samples vary by question type---Simple, Intersection, and Composition---while
ExpertQA (free-form dataset) samples span diverse expert domains. This design captures a broad range of evaluation challenges while enabling a direct comparison between LLM and human judgments.

\subsection{Human Evaluation}
We evaluate reference–generated fact pairs using a structured protocol assessing semantic alignment, factual soundness, and potential evaluation challenges (e.g., ambiguity or underspecification). While core evaluation goals remain consistent across datasets,  guidelines and question formats are tailored to the structure and domain of each dataset.

Annotators are instructed to follow structured, question-driven protocols via written instructions and illustrative examples. Optional contextual resources, such as text content or links to relevant reference material, are provided when appropriate to support informed and interpretable judgments. For the QAMPARI dataset, each annotator dedicated approximately four hours to the task (averaging $\sim$3 minutes per example), and for the ExpertQA dataset, approximately 4.5 hours (averaging $\sim$3.5 minutes per example). All human evaluations were performed by the authors themselves as part of the research process; no external annotators were recruited and no financial compensation was involved. For transparency and reproducibility, the complete set of evaluation protocols, including dataset-specific rubrics, is available in our code repository.

\subsection{LLM Evaluation}
To closely mirror human annotation, we follow an LLM-as-a-Judge method to guide an LLM to assign a label for the reference-generated fact pairs. We employ three well-known LLMs: Claude-3.5-Sonnet~\citep{claude}, Qwen-2.5-32B~\citep{yang2024qwen2}, and LLaMA-3.3-70B-Instruct~\citep{llama3}, chosen based on their reported reasoning and factuality strengths. To minimize inadvertent biases, Qwen-2.5-32B and LLaMA-3.3-70B-Instruct are not used elsewhere in our pipeline, while Claude-3.5-Sonnet is only used in ExpertQA answer generation.

The results, presented in Table~\ref{tab:agreement_analysis}, indicate strong agreement between LLM judgments and human judgments across both datasets. Specifically, LLaMA-3.3-70B-Instruct achieves the highest individual model agreement scores (0.87 for QAMPARI and 0.82 for ExpertQA). This performance is comparable to the LLM majority ensemble (0.87 for QAMPARI and 0.85 for ExpertQA), demonstrating that LLaMA-3.3-70B-Instruct serves as a reliable standalone evaluator. We thus adopt it for our experiments.

\begin{table}[t]
  \centering
  \begin{tabular}{l c c}
    \toprule
    % Comparison                       & Fleiss' $\kappa$ \\ 
    Comparison                       & QAMPARI & ExpertQA \\
    \midrule
    Human Annotators                 & 0.83          & 0.85 \\
    LLM Models                       & 0.72          & 0.70 \\
    Human Maj. vs. LLM Maj.          & \textbf{0.87} & \textbf{0.85} \\
    Human Maj. vs. Claude-3.5-Sonnet & 0.75          & 0.75 \\
    Human Maj. vs. Qwen-2.5-32B      & 0.72          & 0.77 \\
    Human Maj. vs. LLaMA-3.3-70B     & \textbf{0.87} & \textbf{0.82} \\
    \bottomrule
  \end{tabular}
  \caption{\label{tab:agreement_analysis}
     Fleiss' $\kappa$ agreement scores between human annotators and LLM evaluators. The first two rows show within-group agreement; remaining rows show human majority vs. individual LLMs. Scores above 0.80 typically denote substantial agreement.}
    
\end{table}

\begin{table*}[t]
  \centering
  \begin{tabular}{lccc}
    \toprule
    \textbf{Match Type} & \textbf{QAMPARI} & \textbf{RoMQA} & \textbf{ExpertQA}\\
    \midrule
    Full Agreement & 689 & 312 & 11\\
    Partial Agreement & 349 & 376 & 597\\
    ARecall/ERecall Singleton hits & 76 & 42 & 35\\
    Holistic-Prompt$_\text{Gemini}$ Singleton hits & 1,210 & 1,461 & 923\\
    LongRecall$_\text{Gemini}$ Singleton hits & 45 & 95 & 844\\
    LongRecall$_\text{Qwen}$ Singleton hits & 4 & 3 & 320\\
    No Match & 3,314 & 3,126 & 294\\
    \hline
    Total Reference Facts & 5,687 & 5,415 & 3,024 \\
    \bottomrule
  \end{tabular}
  \caption{\label{tab:gt_eval_stats} 
  Counts of match types---full agreement, partial agreement, singleton hits per method, and no match---per dataset. ExpertQA shows high method disagreement (few full agreements), while QAMPARI and RoMQA exhibit more consensus.}
\end{table*}

\subsection{Disagreement Analysis}
To better understand the sources of disagreement between human and LLM-based judgments, we analyzed cases where the majority votes of the two groups diverged across both QAMPARI and ExpertQA datasets. Our analysis revealed that these mismatches arose not from systematic errors but from subtle interpretive differences that reflect fundamental evaluation ambiguities inherent to the task. Below, we describe how these disagreement patterns manifest in each dataset.

\subsubsection*{QAMPARI Disagreement Categories}
QAMPARI disagreements, centered on factual precision and entity boundaries, fall into three categories: \textbf{(I) Missing Specifics}: This pattern reflects disagreement over whether secondary factual details are required for entailment. One example involved a generated answer about \textit{Pittston} that accurately described the town's location but omitted mention of ``Luzerne County,'' a detail explicitly present in the reference. Human annotators judged the description ``match'', while LLMs marked it ``no match'' due to the missing administrative reference. \textbf{(II) Overly Broad Summary}: This pattern centers on whether broad topical relevance is enough, or if precise correspondence is mandatory. In another case, a response about the production of Billie Eilish’s ``\textit{Bad Guy}'' focused on Finneas O’Connell’s general role across her music videos without directly naming his producer credit for the specific song. Although humans accepted this as topically aligned, LLMs rejected it for failing to capture the exact factual target. \textbf{(III) Context Boundary Dispute}: These cases expose differences in how strictly evaluators interpret referential boundaries. A generated answer about \textit{Apui} described the municipality, whereas the reference referred to the ``Apui Mosaic'' conservation area—a related but distinct entity. Here, human annotators judged this as a failure to match, whereas some LLMs accepted it, likely because the lengthy justification in the answer drew focus away from the entity difference.

\subsubsection*{ExpertQA Disagreement Categories}
ExpertQA disagreements, reflecting expert-domain complexity and conceptual completeness, exhibit three distinct categories: \textbf{(I) Scope and Focus Alignment:} Disagreements arose over appropriate coverage scope when generated answers expanded beyond reference boundaries. One economic analysis case involved a reference focused on ``investment inflows and job creation'' while the generated answer discussed broader impacts like ``market access loss and economic uncertainty.'' Human evaluators accepted this comprehensive economic perspective, while LLM evaluators found it misaligned with the reference's specific scope. \textbf{(II) Completeness Threshold Disputes:} Expert topics often involve multi-faceted concepts where partial coverage creates evaluation ambiguity. As another example, a professional ethics case requires both ``communicating value conflicts'' and ``explaining basis of ideals.'' The generated answer thoroughly addressed ethical communication but omitted discussion of foundational principles. Human annotators accepted this partial coverage as conceptually adequate, while LLMs required complete coverage of both specified elements. \textbf{(III) Domain Expertise Interpretation:} This category captures disagreements over whether conceptual understanding of professional principles suffices without direct technical linkages, creating judgment ambiguity where expert knowledge is expressed through different framings. In one legal case, the reference stated that certain workers ``may not be able to claim compensation for dismissal'' due to illegal status, while the generated answer explained that ``the Labour Relations Act applies to legal employment only.'' Both conveyed the same underlying principle, yet LLMs' grasp of legal domain knowledge enabled them to accept this conceptual match, while human evaluators sought more explicit alignment.

The identified disagreement categories reveal nuanced evaluation challenges across domains. Clear evaluation patterns emerged: LLMs demonstrated strict adherence to annotation instructions and followed requirements accurately, while human evaluators exhibited greater flexibility and incorporated contextual knowledge when considering matches. As a result, most disagreement cases showed humans accepting matches that LLMs rejected.

\begin{table*}[t]
    \centering
    \begin{tabular}{c|c|ccc|ccc|ccc}
    \toprule
    \multirow{2}{*}{\textbf{Category}} 
     & \multirow{2}{*}{\textbf{Model}} 
     & \multicolumn{3}{c|}{\textbf{QAMPARI}} 
     & \multicolumn{3}{c|}{\textbf{RoMQA}}
     & \multicolumn{3}{c}{\textbf{ExpertQA}} \\
    \cmidrule(lr){3-5}\cmidrule(lr){6-8}\cmidrule(lr){9-11}
     & & \textbf{P} & \textbf{R} & \textbf{F1} 
       & \textbf{P} & \textbf{R} & \textbf{F1} 
       & \textbf{P} & \textbf{R} & \textbf{F1} \\
    \midrule
    
    \multirow{4}{*}{\shortstack{\textbf{Challenging}\\ \textbf{Subset}\\ (300 cases)}}
     & ARecall/ERecall & 
     0.24 & 0.31 & 0.27 & 
     0.27  & 0.18  & 0.22 &
     0.02  & 0.01 & 0.01    \\
     
     & Holistic-Prompt$_\text{Gemini}$ & 
     0.23 & 0.38 & 0.28 &
     0.30  &  0.52  &  0.38 &
     0.50 & 0.46 & 0.48 \\
     
     & LongRecall$_\text{Gemini}$ &
     0.52 & \textbf{0.86} & 0.65 &
     0.50  &  \textbf{0.92}   &   0.65 &
     0.59 & \textbf{0.69} & 0.63 \\
     
     & LongRecall$_\text{Qwen}$   &
     \textbf{0.84} & 0.70 & \textbf{0.76} &
     \textbf{0.70}  &  0.78   &  \textbf{ 0.74} &
    \textbf{ 0.64} & 0.64 & \textbf{0.64} \\

    \midrule
    
    \multirow{4}{*}{\shortstack{\textbf{Standard}\\ \textbf{Subset}\\ (900 cases)}}
     & ARecall/ERecall                      &
     0.76 & 0.80 & 0.78 &
     0.77 & 0.58 & 0.66 & 
     0.10 & 0.01 & 0.01 \\
     
     & Holistic-Prompt$_\text{Gemini}$      &
     0.36 & 0.92 & 0.52 &
     0.23 & 0.85 & 0.36 &
    0.43  & 0.40 &   0.42   \\
     & LongRecall$_\text{Gemini}$ &
     0.78 & \textbf{0.96} & 0.86 &
     0.69 &   \textbf{0.94}   &   0.80 &
    0.52  & \textbf{0.54}   &   0.53   \\
     & LongRecall$_\text{Qwen}$   &
     \textbf{0.90} & 0.91 & \textbf{0.91} &
     \textbf{0.87}  &  0.89   &   \textbf{0.88} &
    \textbf{0.74}  &  0.44   &   \textbf{0.55}  \\
    
    \bottomrule
    \end{tabular}
    \caption{\label{tab:performance_comparison}
    Precision (P), recall (R), and F1 scores for 5 recall estimation methods on QAMPARI, RoMQA, and ExpertQA, evaluated on reference-generated fact pairs across two sampling subsets: challenging and standard. Note that ARecall is shown for QAMPARI and RoMQA, and ERecall for ExpertQA.}
    
    \end{table*}

\section{Results}

Table \ref{tab:gt_eval_stats} summarizes the distribution of reference–generated fact pairs across three agreement tiers for our evaluation of recall estimation methods: ARecall, ERecall, Holistic-Prompt, and two variants of our proposed LongRecall (using Gemini-1.5-Flash and Qwen-2.5-32B in the entailment checking stage) on three long-form QA benchmarks: QAMPARI, RoMQA, and ExpertQA.
The agreement tiers are defined as follows: (i) \emph{singleton hits}, where a pair is marked as a match by only one method; (ii) \emph{partial agreement}, where more than one method, but not all, mark the pair as a match; and 
(iii) \emph{full agreement}, where all methods mark the pair as a match. 

Due to computational constraints, we conduct evaluation on sampled subsets of the test data. Following established practices in evaluation methodology, we create two complementary evaluation sets to ensure a comprehensive assessment across varying difficulty levels.

The \textbf{standard evaluation set} comprises 900 cases sampled using stratified random sampling to preserve the natural distribution of method agreement patterns observed in the full dataset. Specifically, we maintain proportional representation of full-agreement cases, partial-agreement cases, and singleton-hit cases, ensuring the evaluation reflects the inherent characteristics of each benchmark. This approach follows standard evaluation protocols that maintain representativeness while reducing computational overhead.

The \textbf{challenging evaluation set} consists of 300 carefully selected cases designed to stress-test method robustness on difficult instances where methods typically show disagreement. This subset includes 150 cases where methods show partial agreement and 150 cases representing singleton hits from different methods. To ensure balanced representation across all methods in the singleton category, we employ a quota-based sampling approach that addresses the natural imbalance in method-specific singleton hits by assigning equal representation quotas per method, preventing any single method from dominating the evaluation. This challenging set enables focused analysis of method performance under adversarial conditions, a common practice in robust evaluation frameworks.

\subsection{Comparison with Baselines}

\subsubsection*{LongRecall vs.\ lexical baselines:}
% \paragraph{LongRecall vs.\ lexical baselines.}
Across all three benchmarks, LongRecall consistently outperforms the lexical baselines---ARecall on QAMPARI and RoMQA and ERecall on ExpertQA---by large margins, as reported in Table~\ref{tab:performance_comparison}.
On ExpertQA, lexical methods almost fail as ERecall achieves F$_1$ of merely 0.01, whereas LongRecall$_\text{Gemini}$ scores 0.53 and 0.63 on the standard and challenging splits, respectively.  
Similarly, on both split-span datasets, ARecall scores high on the standard split (0.66 on RoMQA and 0.78 on QAMPARI) because many entity names appear verbatim in the anchors. Nonetheless, it collapses on the challenging split, where exact‐match entity names are rare. 
LongRecall$_\text{Qwen}$, by contrast, achieves F$_1$ of 0.88 (+0.22$\uparrow$) and 0.91 (+0.13$\uparrow$) on the standard subset for RoMQA and QAMPARI, respectively. On the challenging subset, LongRecall$_\text{Qwen}$ widens the gap with ARecall even further by scoring $\sim$0.75 on both datasets, i.e. +0.52$\uparrow$ on RoMQA and +0.49$\uparrow$ on QAMPARI.

ARecall and ERecall suffer from two key weaknesses. First, they yield high false negatives when facts are expressed differently. Second, they incur false positives by matching surface tokens without verifying contextual support. LongRecall overcomes both issues through its candidate selection and entailment checking stages. The semantic filtering step recovers non‐verbatim matches that lexical methods miss, while the entailment check discards unsupported alignments, thus reducing both false negatives and false positives.

\begin{table*}[t]
  \centering
  \begin{tabular}{l|cccc|cccc|cccc}
  \toprule
  \multirow{2}{*}{\textbf{Model}} 
    & \multicolumn{4}{c|}{\textbf{QAMPARI}} 
    & \multicolumn{4}{c|}{\textbf{RoMQA}}
    & \multicolumn{4}{c}{\textbf{ExpertQA}} \\
  \cmidrule(lr){2-5}\cmidrule(lr){6-9}\cmidrule(lr){10-13}
    & \textbf{FP} & \textbf{ FN} & \textbf{FN\textsubscript{CS}} & \textbf{FN\textsubscript{EC}} 
    & \textbf{FP} & \textbf{ FN} & \textbf{FN\textsubscript{CS}} & \textbf{FN\textsubscript{EC}} 
    & \textbf{FP} & \textbf{ FN} & \textbf{FN\textsubscript{CS}} & \textbf{FN\textsubscript{EC}} \\
  \midrule
  LongRecall\textsubscript{Gemini} 
    & 28\% & 4\% & 4\% & 0\% 
    & 28\% & 2\% & 2\% & 0\% 
    & 28\% & 10\% & 4\% & 6\%  \\
  LongRecall\textsubscript{Qwen}   
    & 6\% & 8\% & 4\% & 4\% 
    & 10\% & 8\% & 2\% & 6\% 
    & 14\% & 16\% & 4\% & 12\% \\
  \bottomrule
  \end{tabular}
  \caption{Error analysis comparing LongRecall performance across QAMPARI, RoMQA, and ExpertQA (50 samples each). 
  Values represent percentages of each error type within the 50 samples. FP shows false positive rates; FN shows overall false negative rates, with stage-wise division into FN\textsubscript{CS} (candidate selection misses) and FN\textsubscript{EC} (entailment checking misses).}
  \label{tab:confusion-matrix-combined}
\end{table*}

\subsubsection*{LongRecall vs.\ Holistic-Prompt:}
In a direct comparison, using the same LLM, LongRecall$_\text{Gemini}$ consistently outperforms Holistic-Prompt$_\text{Gemini}$, as shown in Table~\ref{tab:performance_comparison}.
On QAMPARI, LongRecall$_\text{Gemini}$ surpasses the baseline by +0.34$\uparrow$ (F1: 0.86) and +0.37$\uparrow$ (F1: 0.65) on standard and challenging subsets.    
A similar pattern holds for RoMQA. On ExpertQA, LongRecall is superior by +0.11$\uparrow$ and +0.15$\uparrow$ in F1 score on standard and challenging subsets, respectively. 
These results confirm that the LongRecall’s structured candidate selection and entailment checks empowers the same LLM to be markedly better at measuring recall.

LLM-as-a-Judge baselines tackle the mapping task in a single, monolithic prompt that lists \emph{every} reference fact alongside \emph{every} generated fact.  
In this overly long context, the model often (i) links a reference fact to an unrelated generated fact that merely shares surface cues, and (ii) on split-span datasets, maps the reference fact to itself (\textit{self-mapping}) instead of locating the correct generated counterpart.  
These issues manifest clearly in Table~\ref{tab:gt_eval_stats}, where Holistic-Prompt$_\text{Gemini}$ generates exceptionally high singleton-hit counts---on the order of a thousand or more---across all datasets. While self-mapping predominantly affects split-span datasets (driving precision below 0.36), ExpertQA exhibits a different but equally problematic pattern---spurious topical matches where the model incorrectly aligns conceptually related but factually distinct content due to context overload. Notably, ExpertQA shows the highest rate of partial agreements, reflecting the complex semantic relationships that confound holistic evaluation in expert domains. LongRecall dramatically reduces singleton hits across all datasets, demonstrating its effectiveness in handling both self-mapping and spurious semantic alignment issues. 

Using Qwen-2.5-32B in LongRecall improves F$_1$ even further, compared to Gemini-1.5-Flash, thereby demonstrating that LongRecall scales smoothly with entailment verifier quality. Notably, both LLMs show more modest improvements on ExpertQA compared to split-span datasets, suggesting that expert-level domain knowledge remains challenging even for stronger models. This modular design allows easy integration of different LLMs while maintaining the structured approach that addresses context overload---a fundamental limitation of holistic prompting strategies.

\subsection{Error Analysis}
To evaluate the robustness and identify potential failure modes of our LongRecall pipeline, we conducted a component-wise error analysis of fact extraction, candidate selection, and entailment checking using 50 random samples from challenging subsets of QAMPARI, RoMQA, and ExpertQA. This analysis highlights the effectiveness of each stage while identifying specific scenarios where targeted improvements could enhance overall performance across datasets and model variants.

Our error categorization follows the pipeline's sequential structure: false negatives can arise from shortcomings in either candidate selection, where relevant and entailing facts are overlooked and never considered, or from entailment checking, where valid candidates are incorrectly rejected. We therefore divide false negatives into FN\textsubscript{CS} (candidate selection misses) and FN\textsubscript{EC} (entailment checking misses) to isolate stage-specific failures. Conversely, all false positives stem exclusively from entailment checking shortcomings, as any weakly related candidates selected by candidate selection should be appropriately rejected during entailment verification.

\textbf{Fact Extraction}: We manually reviewed the extracted facts to validate the reliability of our rule-based segmentation heuristics and coreference resolution procedure. Across all observed samples, the extracted facts were self-contained and successfully resolved anaphoric expressions to include explicit mentions, eliminating cross-sentence dependencies as intended. We found no evidence of inconsistent granularity between reference and generated fact segmentation, nor structural mismatches that would impede downstream processing. Within the examined samples, the fact extraction component did not exhibit issues that hindered subsequent pipeline steps.

\textbf{Candidate Selection}: Analysis of candidate selection failures reveals distinct dataset-specific challenges, with failure rates varying across domains (2-4\% per dataset) but exhibiting consistent patterns within each dataset regardless of model variant. 
QAMPARI failures (4\% each) stem primarily from focus misalignment where different focus in anchor parts results in minimal lexical similarity, while generated answers' verbosity contrasts sharply with short, concise references, reducing embedding similarity through signal dilution. 
RoMQA demonstrated the lowest candidate selection failure rates (2\% each), but revealed domain-specific challenges with scientific terminology variations and reference fragmentation, where lexical filtering struggles with alternative nomenclature while semantic filtering faces degraded embedding quality from disconnected reference structures. 
ExpertQA exhibited challenges related to expert-level terminology variations, where less common professional expressions conveying the same or closely related meanings are expressed in different forms, presenting difficulties for current filtering approaches that rely on lexical or embedding similarity and demonstrating the nuanced nature of professional discourse in specialized domains.

\textbf{Entailment Checking}: Analysis reveals fundamental model differences in precision-recall trade-offs, with entailment checking serving as the exclusive source of precision errors. Gemini demonstrates consistently permissive behavior, generating 28\% FPs across all datasets by accepting semantically related but insufficient matches through three key patterns: missing temporal specificity (accepting general terms when specific years required), hierarchy confusion mixing specific asnwers with general categories, and technical detail insufficiency accepting general responses when specific requirements are needed. Conversely, Qwen exhibits stricter criteria with substantially lower FP rates of 6\% in QAMPARI, 10\% in RoMQA, and 14\% in ExpertQA. 

False negative analysis reveals Qwen's precision comes at significant recall cost, producing higher FN rates of 4\% in QAMPARI, 6\% in RoMQA, and 12\% in ExpertQA, while Gemini achieved 0\% FNs in QAMPARI and RoMQA, and only 6\% in ExpertQA. Qwen’s failures stem from rigid handling of domain-specific terminology, rejection of valid paraphrases conveying the same meaning, and penalization of additional yet relevant context. Dataset-specific challenges varied across domains: ExpertQA was dominated by professional terminology equivalences and strict domain precision requirements; QAMPARI by tensions between entity precision, generality, and temporal specificity; and RoMQA by variability in terminology usage.

The LongRecall pipeline is generally robust, with accurate Fact Extraction and stable Candidate Selection, though occasional misses (2–4\%) reveal difficulty capturing domain-specific terms and paraphrases. Most performance variation stems from entailment checking, where Gemini favors recall at the cost of more false positives, and Qwen favors precision but misses valid entailments. Future improvements should focus on enhancing the candidate selection’s robustness to terminological and paraphrastic variation, and fine-tuning the entailment checking models to better suit specific domain requirements and desired precision-recall balances.

\section{Conclusion}

We have introduced \textbf{LongRecall}, a foundational framework that addresses a critical gap in measuring the completeness of long-form text generation. By uniquely combining accuracy, interpretability, and scalability in a modular architecture, LongRecall bridges fundamental limitations in existing evaluation methodologies that rely on either shallow lexical matching or opaque holistic judgment. Extensive experiments on diverse QA benchmarks demonstrate that LongRecall consistently and substantially outperforms all baselines, minimizing both false positives and false negatives. By breaking down recall assessment into manageable and interpretable steps, our framework empowers LLMs to achieve markedly higher fidelity in recall estimation.

While the grand challenge of evaluating completeness in fully open-ended settings remains on the horizon, our structured pipeline offers a generalizable building block for diverse evaluation contexts. By establishing robust methodological foundations for recall evaluation, LongRecall supports the development of more complete, reliable, and trustworthy AI systems across a wide range of applications. Future research may build upon this work by integrating domain knowledge, extending to multi-turn and multi-hop scenarios, and developing specialized modules for emerging answer generation paradigms.

\section*{Acknowledgments}
 This research was supported by the Natural Sciences and Engineering Research Council of Canada (NSERC).

% \pagebreak

\bibliographystyle{ACM-Reference-Format}
\bibliography{main}

% \pagebreak

\end{document}